\documentclass[runningheads]{llncs}
\usepackage[T1]{fontenc}
\usepackage{graphicx}
\usepackage{import}
\usepackage{amsmath}
\usepackage{booktabs}
\usepackage{siunitx}
\usepackage{multirow}
\usepackage{color}

\definecolor{myblue}{rgb}{0.21,0.49,0.74}
\usepackage[pagebackref,breaklinks,colorlinks,allcolors=myblue]{hyperref}

\newcommand{\white}[1]{\textcolor[rgb]{1.0,1.0,1.0}{#1}}  
\newcommand{\tb}[1]{\textcolor[rgb]{0.0,0.0,0.0}{#1}}    

\begin{document}

\title{\textit{Deep} in the Jungle:\\Towards Automating Chimpanzee Population Estimation}
\titlerunning{Towards Automating Chimpanzee Population Estimation}

\author{Tom Raynes\inst{1} \and
Otto Brookes\inst{1,3}\orcidID{0000-0001-6865-1844} \and
Timm Haucke\inst{2}\orcidID{0000-0003-1696-6937} \and
Lukas Bösch\inst{3}\and
Anne-Sophie Crunchant\inst{3}\orcidID{0000-0002-4277-2055} \and
Hjalmar Kühl\inst{4}\and
Sara Beery\inst{2}\orcidID{0000-0002-2544-1844} \and
Majid Mirmehdi\inst{1}\orcidID{0000-0002-6478-1403} \and \newline
Tilo Burghardt\inst{1}\orcidID{0000-0002-8506-012X}}

\authorrunning{T. Raynes et al.}
%
\institute{University of Bristol, UK\and
 Massachusetts Institute of Technology, USA\and
 Wild Chimpanzee Foundation\and
 Senckenberg Museum of Natural History, Germany\vspace{-8pt}
 }
 
\maketitle              
\begin{abstract}
The estimation of abundance and density in unmarked populations of great apes relies on statistical frameworks that require animal-to-camera distance measurements. In practice, acquiring these distances depends on labour-intensive manual interpretation of animal observations across large camera trap video corpora. This study introduces and evaluates an only sparsely explored alternative: the integration of computer vision–based monocular depth estimation~(MDE) pipelines directly into ecological camera trap workflows for great ape conservation. Using a real-world dataset of 220 camera trap videos documenting a wild chimpanzee population, we combine two MDE models---Dense Prediction Transformers and Depth Anything---with multiple distance sampling strategies. These components are used to generate detection distance estimates, from which population density and abundance are inferred. Comparative analysis against manually derived ground-truth distances shows that calibrated DPT consistently outperforms DepthAnything. This advantage is observed in both distance estimation accuracy and downstream density and abundance inference. Nevertheless, both models exhibit systematic biases. We show that, given complex forest environments, they tend to overestimate detection distances and consequently underestimate density and abundance relative to conventional manual approaches. We further find that failures in animal detection across distance ranges are a primary factor limiting estimation accuracy. Overall, this work provides a case study beyond pure chimp distance estimation demonstrating practically that MDE-driven camera trap distance sampling is a viable alternative to manual distance estimation via a case study. The proposed approach yields density and abundance estimates within $22\%$ of those obtained using traditional methods. Taken together, these results indicate a realistic pathway toward scalable and automated population modelling for great apes.\vspace{-8pt}
\end{abstract}

\section{Introduction}\vspace{-8pt}

\textbf{Motivation --} Ecosystems worldwide are experiencing alarming declines in biodiversity, with primates among the most threatened taxonomic groups globally.
Approximately 60\% of all primate species are classified as threatened with extinction by the IUCN~\cite{Bezanson2019}. In particular, all great ape species and subspecies—including orangutans, gorillas, and chimpanzees (including bonobos)—are listed as Endangered or Critically Endangered, with all but one subspecies now in decline~\cite{Hockings2015}.

Given the critical conservation status of great apes, accurate species-level data—such as population size and density estimates—are essential for informing timely and effective conservation action. Camera traps have become the de facto tool for large-scale great ape population monitoring~\cite{ct_success}. These motion-triggered sensors capture wildlife in natural habitats, enabling data collection at spatial and temporal scales infeasible with traditional survey methods. However, while camera traps reduce field survey costs, downstream analysis remains heavily reliant on labour-intensive manual processing. The long timescales required for expert-driven annotation limit scalability, highlighting the need for resource-efficient monitoring solutions that enable population metrics—such as density and abundance—to be derived rapidly and reliably~\cite{conservation_data}.

\textbf{Estimating Population Statistics --} At large spatial or temporal scales, individual great apes are typically unidentifiable; consequently, population estimation relies on unmarked population methodologies, including the random encounter model~\cite{rem}, the instantaneous estimator~\cite{IS}, and camera trap distance sampling (CTDS)~\cite{howe2017distance}, which is the focus of this study. While these approaches differ in their assumptions, all require estimation of distance-related quantities; for CTDS, this corresponds to animal-to-camera distances. As most camera traps are monocular, they lack direct depth information, making distance estimation non-trivial. Conventional approaches therefore rely on manual comparison of observations with reference material captured at known distances~\cite{HAUCKE2022101536}. This process is time-consuming, difficult to scale, and introduces subjectivity, creating a major bottleneck. Automating distance estimation~\cite{deepchimp} reliably and directly evaluated in the CTDS pipeline would therefore hugely improve monitoring efficiency and reduce turn-around times for conservation action.

\begin{figure}[!h]
    \centering
\includegraphics[width=0.92\textwidth]{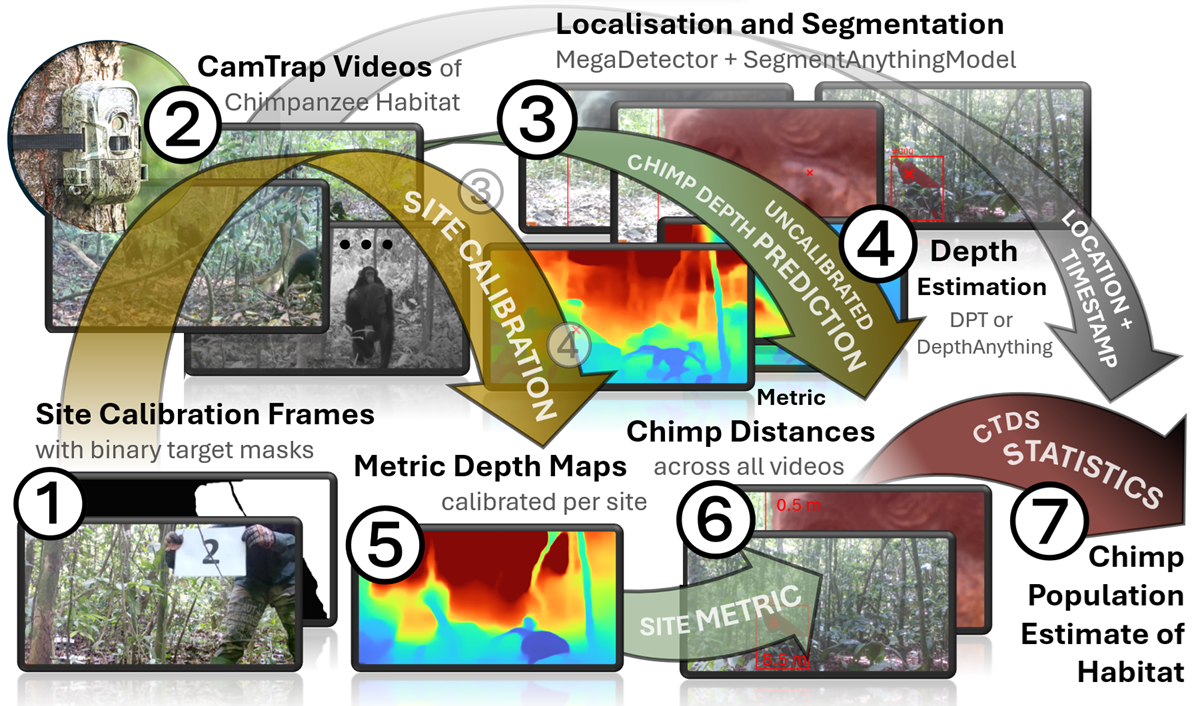}\vspace{-4pt}
\caption{\textbf{Overview of Proposed Approach.} \textbf{{(1)}}~Frames containing calibration markers at known distances, and \textbf{{(2)}}~habitat frames sampled from motion-triggered camera trap videos, are \textbf{{(3)}}~passed to MegaDetector and SAM to localise and segment content of interest. These are then processed via \textbf{{(4)}}~monocular depth estimation, utilising the approach of Haucke et al.~\cite{HAUCKE2022101536}, to yield raw depth maps of observations. The framework further combines \textbf{{(5)}}~calibrated metric depth maps with the raw data to produce \textbf{{(6)}}~distance estimates for all detected chimpanzees. These estimates are combined with camera trap metadata (spatial and temporal) and statistically processed to yield \textbf{{(7)}}~density and abundance estimates using the CTDS statistical framework. Overall, this pipeline streamlines camera trap distance sampling from days or weeks of manual labour to hours of computation.}
\label{fig:abstract_workflow}
\end{figure}

\textbf{Paper Concept --} In this work, we evaluate AI-driven CTDS automation for great ape camera trapping and develop an automated population estimation pipeline applicable to real-world chimpanzee datasets from the West African rainforest. Although monocular depth estimation (MDE) is a promising tool for population monitoring, its application in this context has received limited attention. Building on prior depth estimation work, we assess the feasibility of MDE pipelines directly in conservation-relevant environments, analysing trends in distance, density, and abundance estimates across parameterisable configurations. We evaluate accuracy relative to manual approaches, identify optimal configurations, and highlight limitations requiring future refinement. Critically, we demonstrate in a case study that automated distance estimates can be used effectively to derive local population density and abundance, establishing a practical pipeline from raw camera trap video to population estimates~(Fig.~\ref{fig:abstract_workflow}).

\vspace{-12pt}\section{Related Work}\vspace{-9pt}

{\bf Computer Vision for Ecology --} Recent years have seen rapid progress in the application of computer vision (CV) to ecological and animal studies, driven by the release of large-scale wildlife datasets and advances in specialised animal-focused models. These developments span a wide range of tasks, including species classification~\cite{gu2025bioclip}, animal detection, and tracking~\cite{naik2024bucktales}
In parallel, progress in general-purpose CV—particularly in segmentation (e.g. SAM-based approaches~\cite{wasmuht2025sa}) and monocular depth estimation (MDE; e.g. Depth Anything~\cite{lin2025depth})—has enabled increasingly robust scene understanding from unconstrained footage. Together, these advances have lowered the barriers to deploying CV in ecology. Despite this progress, relatively few studies have examined how such methods translate to real-world ecological applications or evaluated their effectiveness in deriving ecologically meaningful measurements for downstream analysis~\cite{whytock2021robust,pantazis2024deep,christensen2024moving,henrich2024semi}.

{\bf Automating the Generation of Population-level Statistics --}
Within this domain, increasing attention has been given to applying CV models to camera trap data and using their predictions as inputs for estimating downstream ecological statistics.
Whytock et al.~\cite{whytock2021robust} applied a ResNet-50 model to classify 26 Central African mammal and bird species, demonstrating statistical equivalence between model-derived and expert annotations when estimating species richness and occupancy.
Pantazis et al.~\cite{pantazis2024deep} examined the same metrics, showing that dataset scale and annotation quality exert greater influence than architectural choice. Brookes et al.~\cite{brookes2023triple} applied behaviour recognition models to identify and exclude video segments containing behaviours known to bias population estimates, demonstrating that automated filtering can lead to substantial discrepancies between model-derived and expert-derived estimates.

{\bf Monocular Depth Estimation for CTDS --}
Haucke et al.~\cite{HAUCKE2022101536} showed that MDE, supported by reference imagery, can automate animal-to-camera distance estimation for CTDS, reducing manual effort by over an order of magnitude. Johanns et al.~\cite{johanns2022automated} later introduced a fully automated DPT-based pipeline producing metric depth estimates without reference material. Subsequent work demonstrated that semi- and fully automated MDE pipelines yield population density estimates comparable to manual workflows across multiple species, while substantially reducing annotation effort~\cite{henrich2024semi,bak2025automated}. More recent studies have examined joint automation of species classification and distance estimation, revealing non-linear effects of classification accuracy on downstream CTDS outputs~\cite{henrich2025camera}, and have further reduced manual intervention through fully automated mask generation~\cite{automated_masking}. However, downstream CTDS analyses remain underexplored. Building on this work, we systematically evaluate combinations of localisation and MDE models and explicitly quantify their impact on abundance and density estimates for great ape populations.\vspace{-12pt}
\section{Dataset}\vspace{-8pt}
The dataset, provided by the Wild Chimpanzee Foundation (WCF), comprises 220 motion-triggered camera trap videos collected from 65 fixed locations across Taï National Park, Côte d’Ivoire -- one of the largest remaining primary rainforests in West Africa. The videos capture chimpanzees under unconstrained conditions, exhibiting substantial variation in viewpoint, distance, illumination, occlusion, and background clutter. For each camera location, reference videos were recorded; a surveyor appearing at known distances up to 15m, enabling camera-specific depth calibration. Chimpanzee detections were annotated at two-second intervals while individuals were visible, with the first annotation recorded once 50\% of the projected body area was present in the frame. Annotations include temporal metadata and an expert-estimated animal-to-camera distance derived from the corresponding reference videos.\vspace{-12pt}
\section{Experiments}\vspace{-6pt}

\textbf{Computer Vision Pipelines for Chimpanzee Distance Estimation}. We estimated chimpanzee distance from monocular camera‑trap frames using a modular pipeline~(see Fig.~\ref{fig:abstract_workflow}) building on Haucke et al.~\cite{HAUCKE2022101536}. In our work, animals were first detected with MegaDetector~\cite{mega_detector}, yielding bounding boxes that were also refined into instance masks via Segment Anything~\cite{segment_anything}. Monocular depth was then predicted per animal using DPT~\cite{dpt} or Depth Anything~\cite{depth_anything}, and relative depth was converted to metric distance using camera‑specific reference frames containing a human at known distances up to 15\,m. For bounding‑box representations~(BBOX), distance was defined as the 20th percentile of depth values within the box to reduce background influence, whereas for segmentation‑based representations~(SEG) it was taken from the depth at the mask’s geometric centre. During calibration, only animal‑mask pixels were excluded, improving scale alignment at short ranges. Combining two instance representations~(BBOX, SEG) with two depth models~(DPT, Depth Anything) produced four experimental configurations.

\textbf{Evaluation Protocol}. Distance estimation was evaluated under two frame sampling regimes: (\textbf{i})~frames aligned with human-annotated chimpanzee observations for quantitative comparison against manual distance estimates, and (\textbf{ii})~frames sampled uniformly at two-second intervals across all videos, including frames without chimpanzees, to simulate fully automated deployment.

\textbf{Downstream Ecological Analysis}. Model-predicted distances were used as input to a statistical model (i.e., CTDS~\cite{howe2017distance}) to estimate both chimpanzee density and abundance. For annotated frames, model predictions replaced manual distance estimates where applicable, while for automated frame processing model-derived distances were assigned generally. Distance sampling followed established practice~\cite{howe2017distance}, including truncation, binning, and selection of detection functions using adjusted Akaike Information Criterion~(QAIC).\vspace{-8pt}

\subsection*{RESULT 1 - Distance Estimation Benchmarks}
\label{subsec:distance_accuracy}\vspace{-2pt}

\textbf{Comparison to Manual Ground Truth}. We evaluated precision and accuracy of distance estimates produced by the four pipeline configurations by benchmarking model-predictions via mean average error (MAE) and root mean squared error (RMSE) against manually annotated distances provided by WCF with the dataset. Evaluation was performed on detection frames drawn from the manually annotated sample only, enabling direct comparison to expert-derived distance estimates. Table~\ref{tab:overall_errors} reports MAE, RMSE, and mean signed difference ($\Delta_{\text{AVG}}$) for each configuration. While MAE and RMSE quantify absolute error magnitude, $\Delta_{\text{AVG}}$ captures systematic bias, indicating whether a configuration tends to over- or under-predict distance relative to manual estimates.\vspace{-4pt}

\begin{table}[!h]
    \centering
   \caption{\tb{\textbf{Chimpanzee Distance Estimation Performance.} Mean average error~(MAE), root mean squared error~(RMSE), average difference between model and manual estimate ($\Delta_{AVG}$) in meters (m) showing some performance advantages of DPT over DA. These data describe estimates for each pipeline configuration collectively.}}
    \label{tab:overall_errors}
    \begin{tabular}{lccc}
        \textbf{Method \white{.......}} & \textbf{MAE} (m) \white{.} & \textbf{RMSE} (m) \white{.} & \textbf{$\Delta_{AVG}$} (m) \\
        \midrule
        DPT, BBOX & 1.81 & 2.66 & \textbf{0.59} \\
        DPT, SEG  & \textbf{1.70} & \textbf{2.45} & 0.84 \\
        DA, BBOX  & 2.03 & 2.62 & 1.49 \\
        DA, SEG   & 3.00 & 3.52 & 2.80 \\
    \end{tabular}
\end{table}\newpage

\textbf{Single Animal Constraint}. To use all 2,118 potentially useable frames, manual annotations would have to be paired with corresponding model predictions for \textit{each} animal; however, the absence of frame-level identity information for manual annotations introduces ambiguity in scenes containing multiple individuals. In such cases, it is not possible to unambiguously associate a modelled distance with a specific manual annotation without full animal re-identification. To avoid this confound, evaluation was restricted to 1,659 frames containing a single chimpanzee, for which a one-to-one correspondence between manual and modelled distances can be established. For these frames, modelled distance estimates were paired with their corresponding manual estimates and aggregate error statistics were computed.
\begin{figure}[t]
    \centering
\includegraphics[width=1\textwidth]{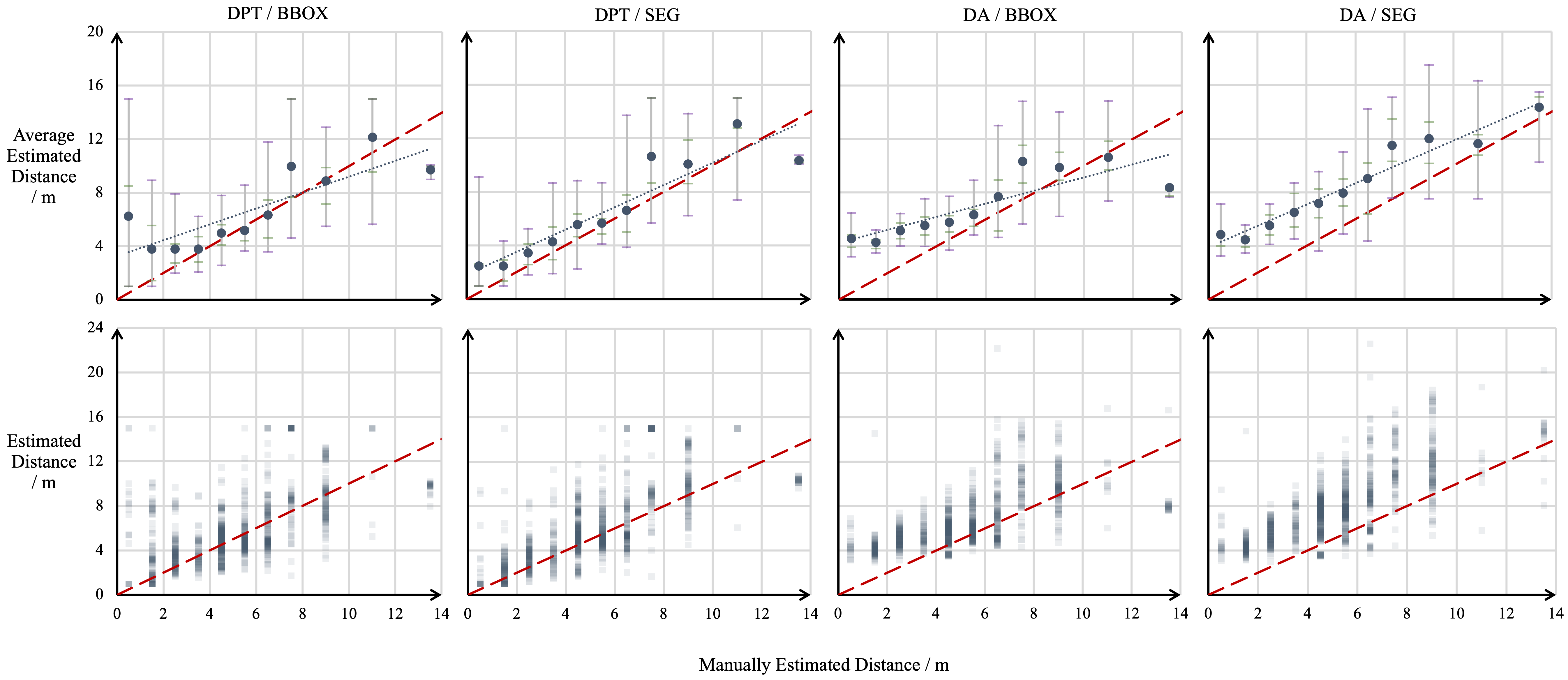}
    \caption{\textbf{Chimpanzee Distance Estimation Performance by Distance.} Graphs showing both individual (bottom) and mean (top) modelled distance estimates mapped to their corresponding manual
        estimates for all configurations. The blue dotted line
        shows the fitted regression line. The red dashed line shows the ideal (i.e., model=manual). The error bars show the 25–75 (green) and 5–95 (purple) percentiles (best viewed zoomed). There is a reasonable correlation between manual and modelled distance, although the level of correlation varies.}
    \label{fig:distance_comparison}
\end{figure}

\begin{figure}[!b]
    \centering
    \includegraphics[width=1.02\textwidth]{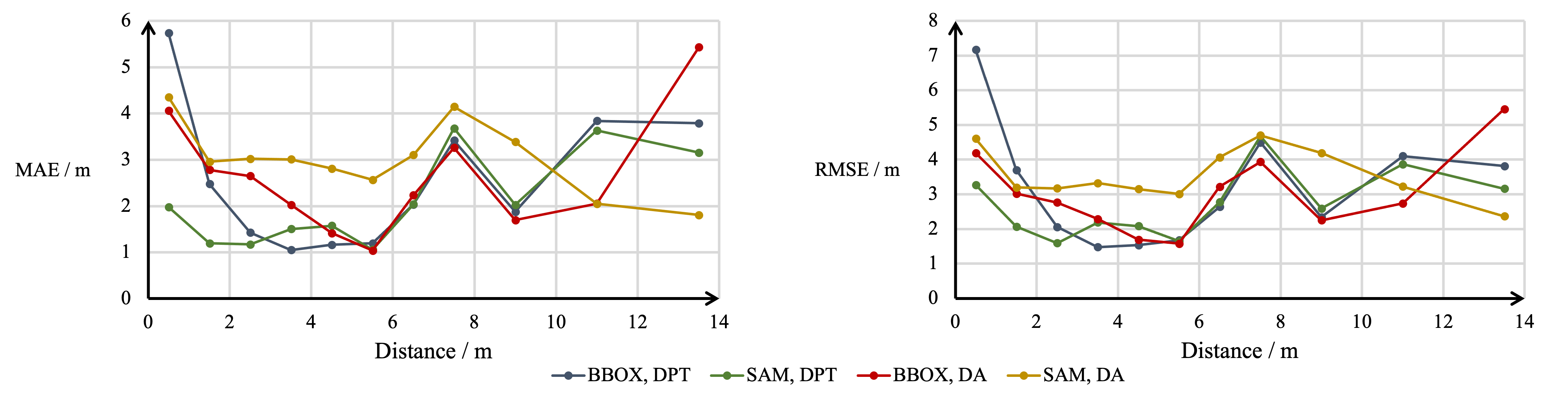}
    \caption{\textbf{Distance Estimation Error by Ground Truth Distance.} Mean average error (left) and root mean squared error (right) for distance estimates
    grouped by their corresponding manual estimates for each pipeline configuration.
   MAE is generally high at the extremes of the distance scale with an additional spike at
    approx. seven meters.}
    \label{fig:error_graphs}
\end{figure}

\textbf{Estimation Performance by Distance}. To analyse performance as a function of distance, modelled estimates were grouped by their corresponding manual distances (0.5\,m, 1.0\,m, $\ldots$, 15\,m). Fig.~\ref{fig:distance_comparison} shows the mean modelled distance for each group and the full distribution of individual estimates. Linear regression was fitted between grouped modelled and manual distances, and the resulting slopes (see Table~\ref{tab:regression_gradients}) quantify how well each configuration captures depth scale. Slopes below one indicate systematic compression of depth scale~(essentially overestimating distances), with segmentation-based configurations exhibiting consistently steeper slopes than their bounding-box counterparts. Finally, per-distance error statistics were computed to examine how estimation accuracy varies across the distance range. Fig.~\ref{fig:error_graphs} shows the MAE and RMSE, respectively, as a function of manual distance, highlighting systematic trends in error across near, mid-range, and distant detections.

\begin{table}[!h]
    \footnotesize
    \centering
    \vspace{-4pt}
  \caption{\tb{\textbf{Systematic Distance Estimation Errors.} Regression slopes relating binned model-predicted distances to corresponding manual distance estimates for each configuration reveal systematic compression of depth scale~(i.e. \textit{all} numbers smaller than 1.00, the value of exact reproduction). Slopes were computed from the regression lines shown in Fig.~\ref{fig:distance_comparison}.}}\vspace{-4pt}
    \label{tab:regression_gradients}
    \begin{tabular}{lc}
        \textbf{Method} & \textbf{Regression Slope}\\
        \midrule
        DPT, BBOX & 0.59 \\
        DPT, SEG  & \textbf{0.84} \\
        DA, BBOX  & 0.49 \\
        DA, SEG   & 0.80 \\
    \end{tabular}\vspace{-7pt}
\end{table}

\subsection*{RESULT 2 - Impact of Segmentation vs. Box Localisation}\vspace{-6pt}
\label{subsec:detection_effects}

\textbf{Evaluating Impact of Localisation Detail}. We next analyse the impact of localisation granularity on distance estimation accuracy, comparing bounding box (BBOX) and segmentation-based (SEG) representations. When paired with DPT, segmentation yields a substantial improvement in both accuracy and precision at close distances ($<2$\,m). For detections corresponding to a manual distance of 0.5\,m, BBOX results in a MAE of 5.74\,m and RMSE of 7.17\,m, whereas SEG reduces error to a MAE of 1.98\,m and RMSE of 3.27\,m, with a markedly narrower interquartile range. This improvement is underpinned by both more appropriate calibration alignment as well as more effective background exclusion, resulting instances where calibration collapses for BBOX under DPT while segmentation yields plausible distance estimates.

\begin{figure}[!h]
    \centering
    \begin{minipage}[t]{0.49\textwidth}
        \centering
        \includegraphics[width=\linewidth]{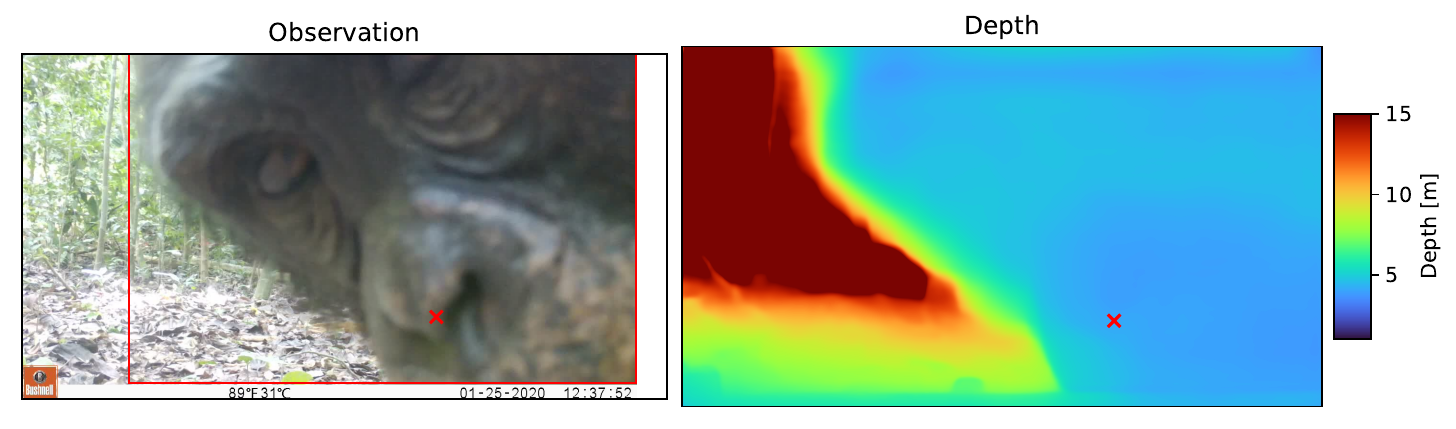}
    \end{minipage}
    \begin{minipage}[t]{0.49\textwidth}
        \centering
        \includegraphics[width=\linewidth]{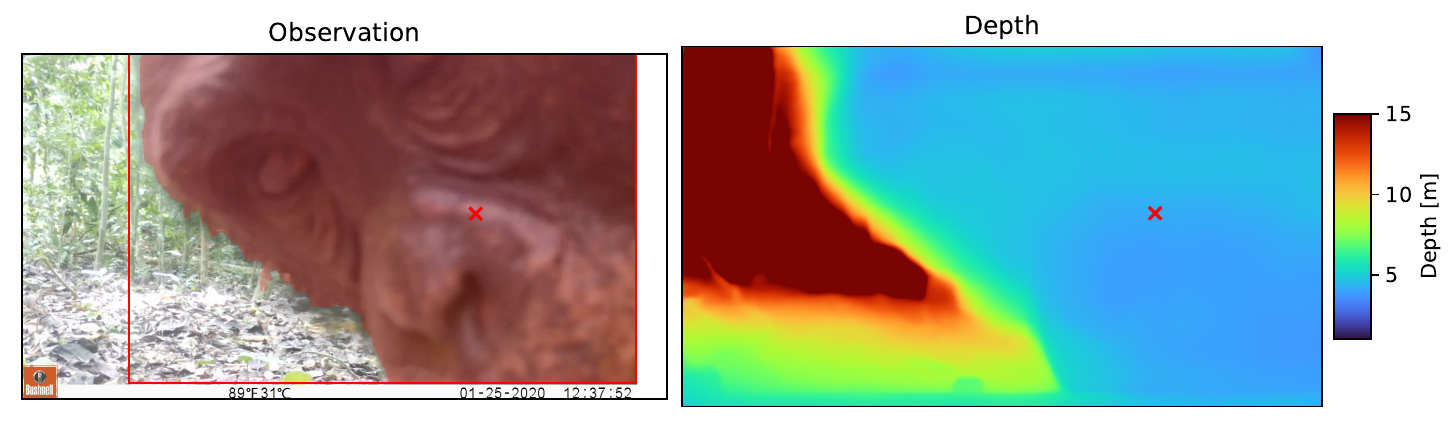}
    \end{minipage}
    \caption{\textbf{ Close-up Distance Estimations.} Example of Depth Anything depth maps generated using bounding box (left) and segmentation
        (right) detection methods at a detection distance of 0.5 meters (manual estimate). In this example, the bounding box method gives a distance estimate of 4.11 meters while the segmentation method gives a distance estimate of 4.17 meters. The distance estimates are close due to identical depth maps, unlike DPT where calibration is applied.}\vspace{-12pt}
    \label{fig:bbox_vs_seg_close_da}
\end{figure}

In contrast, this effect is not observed when using Depth Anything. At close distances, BBOX and SEG produce comparable accuracy and precision (e.g., at 0.5\,m, MAE of 4.06\,m vs.\ 4.35\,m), reflecting the fact that Depth Anything produces metric depth estimates that have no need to be re-scaled using reference frames. Consequently, differences between BBOX and SEG in this setting arise primarily from pixel sampling inaccuracies rather than calibration effects. Fig.~\ref{fig:bbox_vs_seg_close_da} shows that both representations yield identical depth maps, with only minor differences in the final distance estimate.

\textbf{Estimation Performance by Distance}. Across the full distance range, segmentation-based representations consistently produce steeper regression slopes than their bounding-box counterparts (Table~\ref{tab:regression_gradients}), indicating improved preservation of depth scale against the ground truth. This effect is most pronounced at medium to long distances, where partial occlusion becomes more common. Bounding boxes frequently include foreground occluders at shorter distances, biasing percentile-based depth estimates downward, whereas segmentation more effectively isolates the target individual.

\begin{figure}[!b]
    \centering
    \makebox[0.8\textwidth][c]{
        \includegraphics[width=0.8\textwidth]{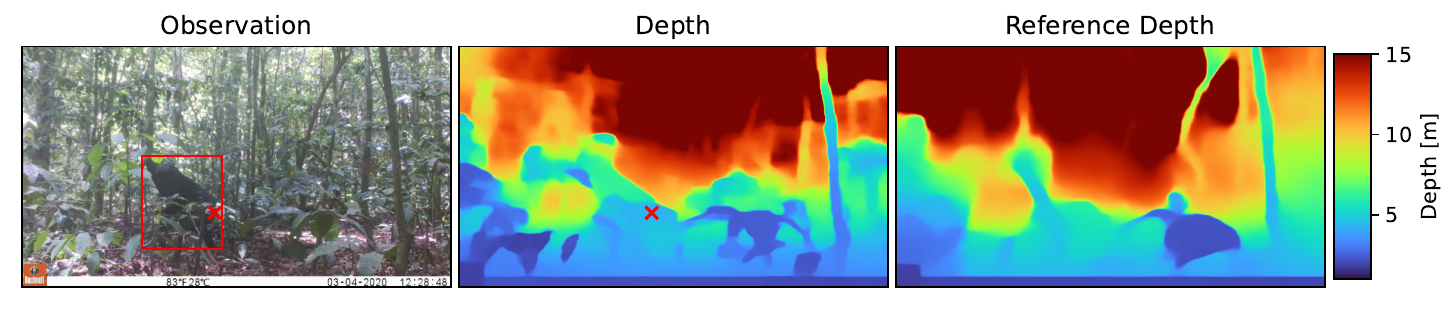}
    }\\[1mm]
    \makebox[0.8\textwidth][c]{
        \includegraphics[width=0.8\textwidth]{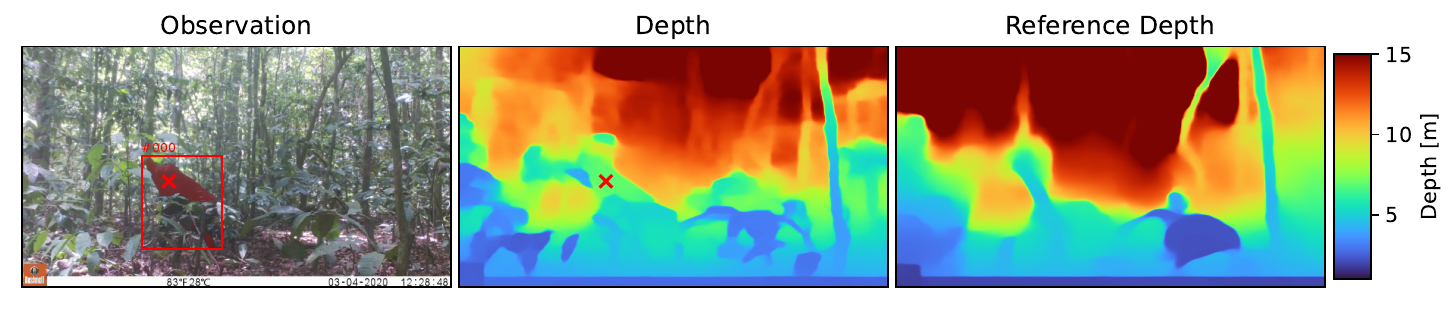}
    }\vspace{-5pt}
    \caption{\textbf{Impact of Occlusions on Bounding Box–based Distance Estimation.} Example DPT depth maps generated using bounding box–based (top) and segmentation-based (bottom) detection methods at a manual detection distance of 6.5m, where the individual is partially occluded by foliage. The bounding box method estimates a distance of 4.40m, while the segmentation method estimates 6.96m.}
    \label{fig:bbox_vs_seg_occluded}
\end{figure}

\begin{figure}[!b]
    \centering
    \makebox[0.9\textwidth][c]{
        \includegraphics[width=0.9\textwidth]{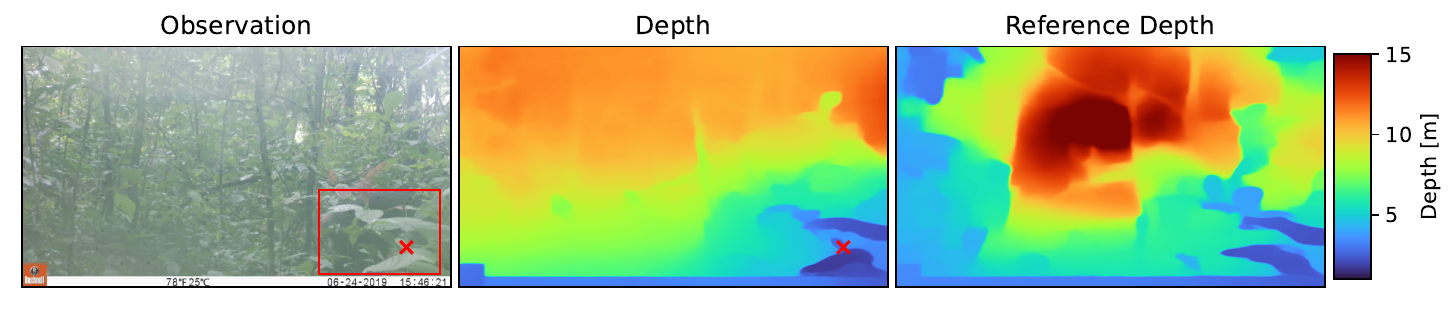}
    }\\[1mm]
    \makebox[0.9\textwidth][c]{
        \includegraphics[width=0.9\textwidth]{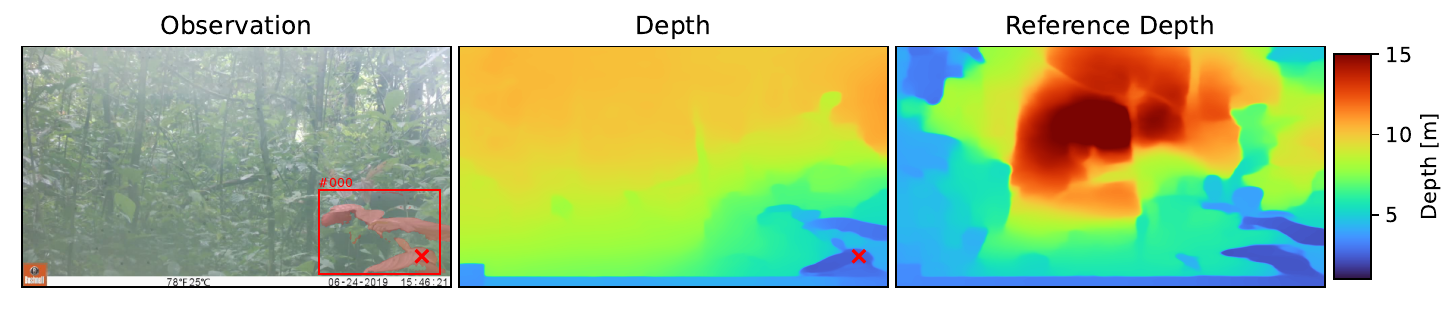}
    }\vspace{-9pt}
    \caption{\textbf{Environmental Conditions impact Segmentation-based Estimation.} Example of DPT depth maps generated using bounding box (top) and segmentation (bottom) detection methods at a detection distance of 4.5m (manual estimate) where the detected individual is partially occluded by foliage. Here, moisture on the camera lens has resulted in a slightly hazy image, leading to a failed segmentation of the detected individual. The bounding box method gave an estimated distance of 2.64m while segmentation gave a distance estimate of 2.56m.}
    \label{fig:haze_occluded}
\end{figure}

Fig.~\ref{fig:bbox_vs_seg_occluded} demonstrates such a case, where BBOX underestimates distance due to occluding foliage, while SEG recovers a value close to the manual estimate. When segmentation quality degrades due to poor image quality~(e.g., haze) or occluding structures are incorrectly separated during segmentation, underestimation nevertheless occurs~(see Fig.~\ref{fig:haze_occluded}). In addition, the steeper regression slope observed for segmentation partially explains the increased absolute error observed for Depth Anything with SEG, where the regression line does not intersect the identity line. In this case, MAE and mean signed error are similar in magnitude, indicating systematic over-prediction across distances.\vspace{-4pt}

\subsection*{RESULT 3 - Effect of Depth Model Architecture}\vspace{-2pt}
\label{subsec:depth_model_effects} 

\textbf{Reliability of Chimpanzee Distance Estimates.} We next analyse the impact of MDE architecture on distance estimation accuracy. As shown before in Table~\ref{tab:regression_gradients}, DPT yields consistently steeper regression slopes than Depth Anything, indicating closer alignment to the true depth scale. Across configurations, DPT slopes are nearer to the ideal value of one, suggesting improved recovery of absolute depth. At short to medium distances (2–7\,m), Figure~\ref{fig:distance_comparison} showed that both depth models capture relative depth structure well, with binned estimates following an approximately linear trend. However, Depth Anything systematically over-predicts distance in this range, whereas DPT estimates more closely follow the identity line. This trend extends to extreme close-range detections for Depth Anything under both detection representations, and to DPT when paired with segmentation, while DPT with bounding boxes exhibits the over-prediction behaviour discussed previously.

\begin{figure}[!t]
    \centering
    \makebox[0.8\textwidth][c]{
        \includegraphics[width=0.7\textwidth]{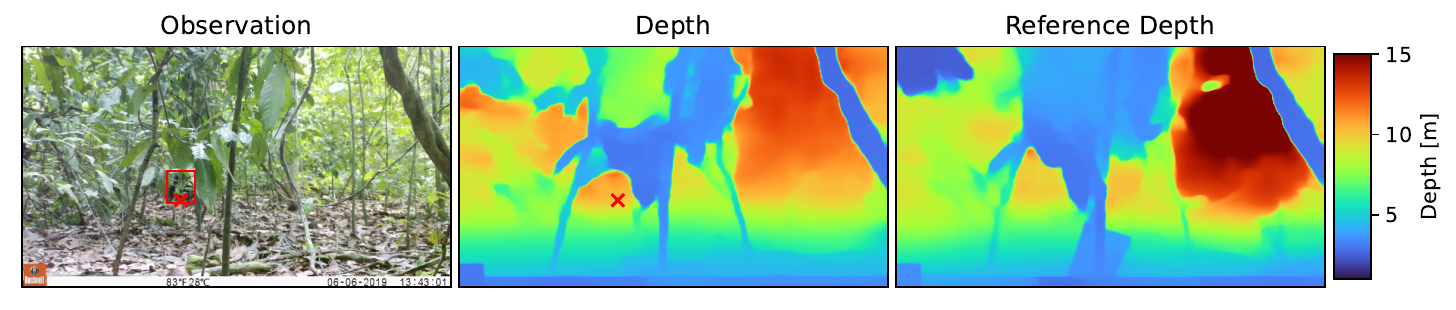}
    }\\[1mm]
    \makebox[0.8\textwidth][c]{
        \includegraphics[width=0.7\textwidth]{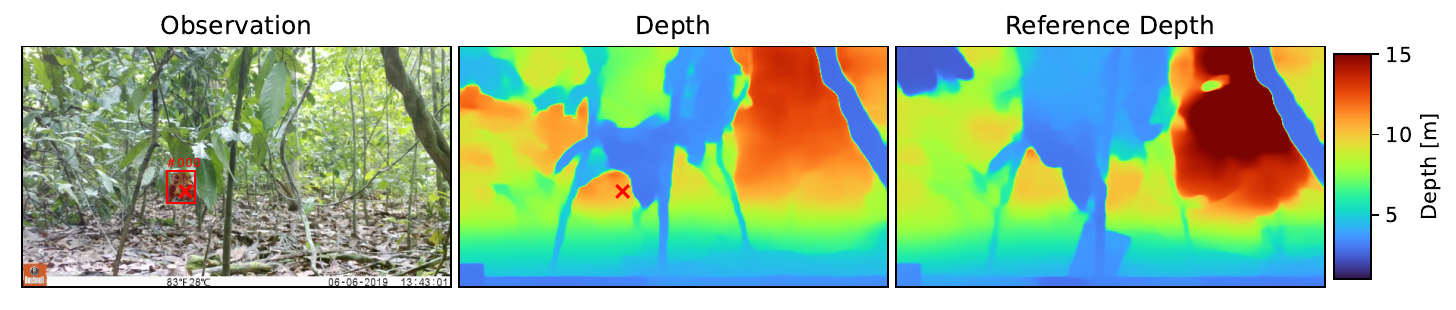}
    }\\[1mm]
    \begin{minipage}[t]{0.49\textwidth}
        \centering
        \includegraphics[width=\linewidth]{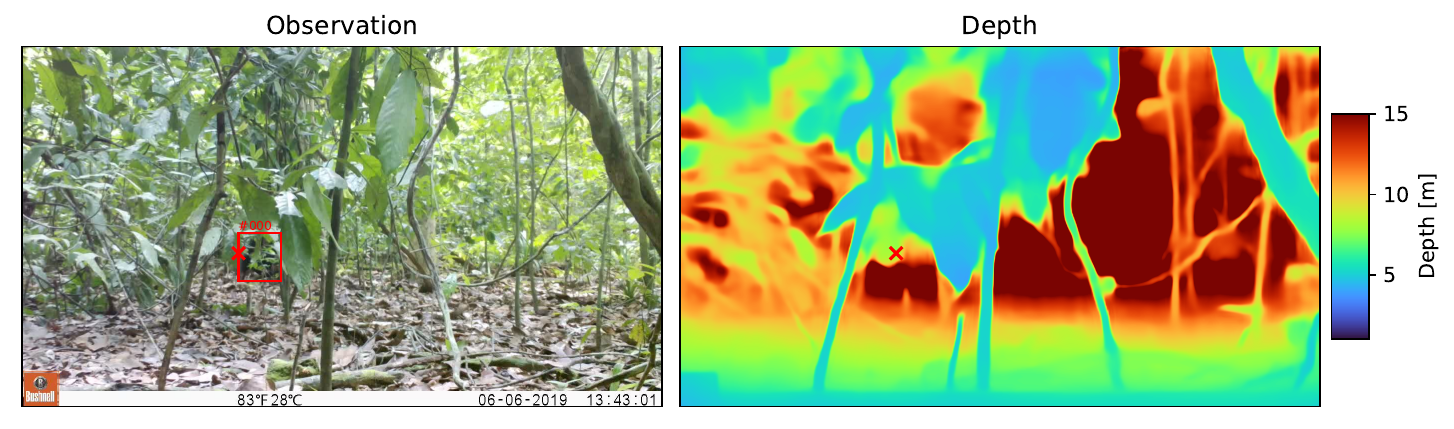}
    \end{minipage}
    \begin{minipage}[t]{0.49\textwidth}
        \centering
        \includegraphics[width=\linewidth]{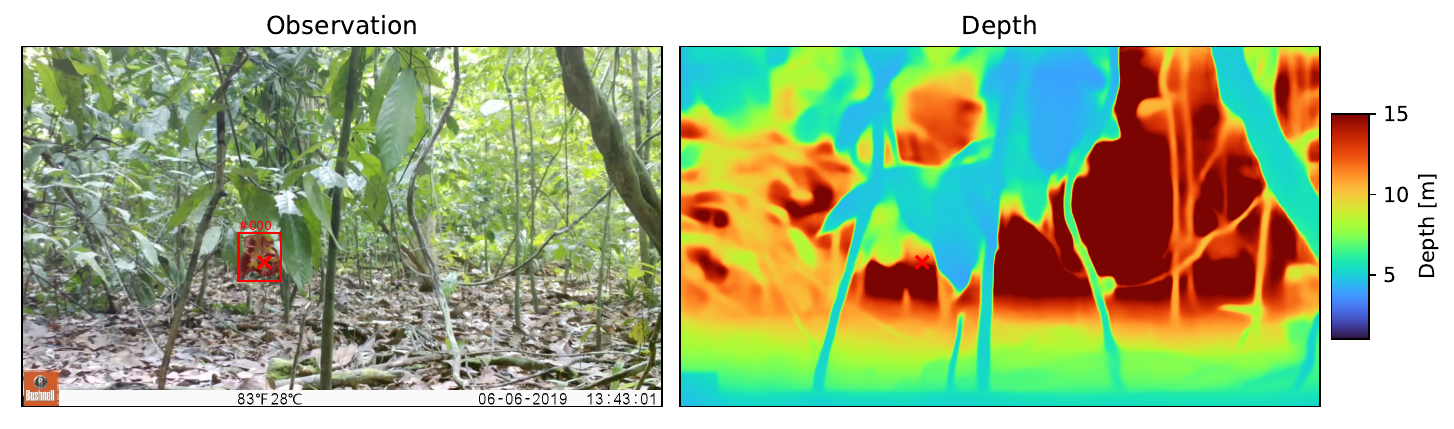}
    \end{minipage}
    \caption{\textbf{Fine-Granularity of Spatial Depth Estimates}. Example of depth maps generated using DPT/BBOX (top), DPT/SEG (middle), DA/BBOX (bottom/left) and DA/SEG (bottom/right)  methods at a detection distance of 13.5m (manual estimate). The following distance estimates were given: DPT/BBOX = 10.0m, DPT/SEG = 10.4m, DA/BBOX = 7.83m, DA/SEG = 13.6m. The depth maps show that Depth Anything is capturing the depth contours of fine structures more vividly than DPT.}
    \label{fig:fine_detail}
\end{figure}

\textbf{Fine-grained Detail from Depth Anything.} Despite its weaker recovery of absolute scale, Depth Anything produces depth maps with substantially finer spatial detail. This is illustrated in Fig.~\ref{fig:fine_detail}, which shows depth predictions for all configurations on a common detection frame. Depth Anything more clearly resolves depth contours within small detection regions, whereas DPT depth maps appear comparatively flattened at longer distances. This difference in depth structure explains the divergent behaviour observed at long range. For detections corresponding to a manual distance of 13.5m, both DPT configurations and Depth Anything with bounding boxes significantly under-predict distance (seen before in Fig.~\ref{fig:distance_comparison}). In contrast, Depth Anything combined with segmentation yields estimates closely aligned with the manual distance. As shown in~Fig.~\ref{fig:fine_detail}, DPT fails to capture sufficient depth variation within the detection region, resulting in similar distance estimates of approximately 10m regardless of representation. Depth Anything, by contrast, resolves fine-grained depth differences that allow pixel sampling strategies to diverge: bounding-box percentile sampling is biased by nearby occluding pixels, while segmentation-based sampling isolates the central instance pixel, producing an accurate estimate of 13.6m.

\begin{figure}[!ht]
    \centering
\hspace{-20pt}\includegraphics[width=0.95\textwidth]{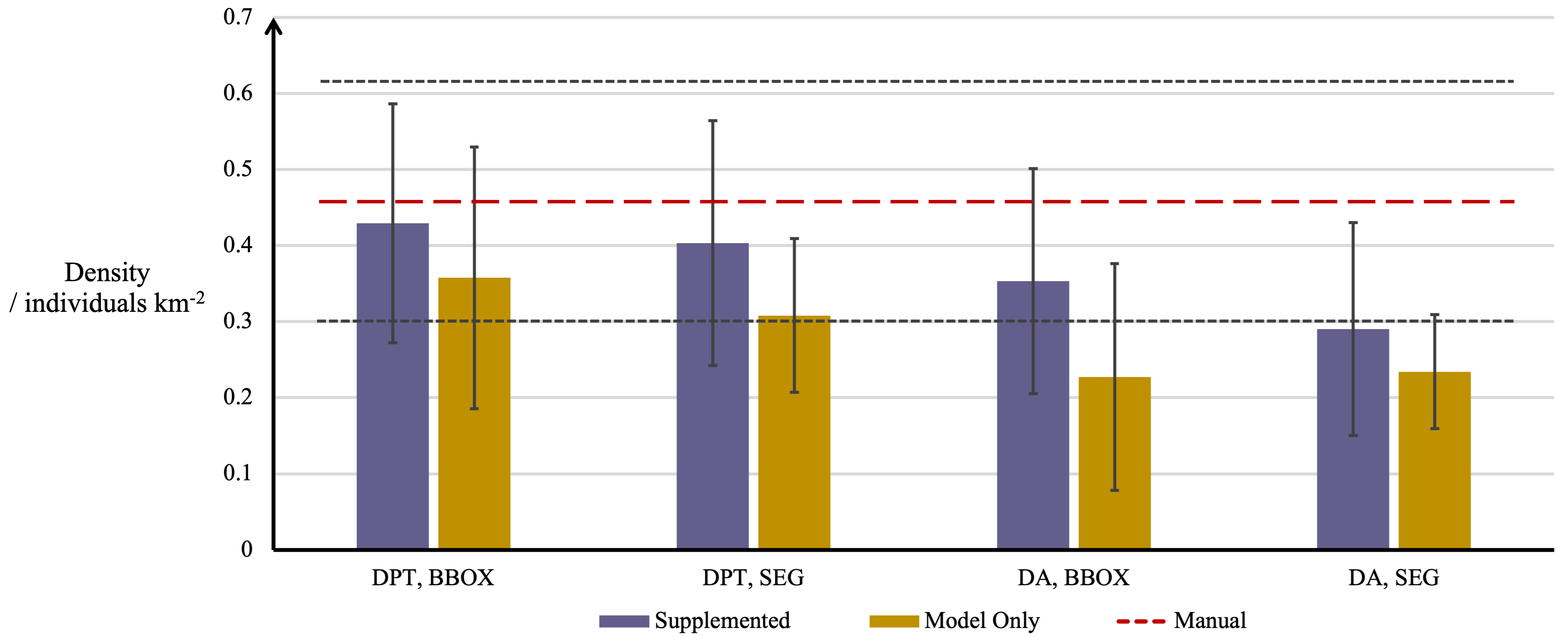}
    \caption{\textbf{Full Pipeline Evaluation: Population Estimation Case Study.} Density estimates for the chimpanzee population in Taï National Park, derived from 220 camera trap videos from the WCF dataset. Estimates are shown for each distance estimation pipeline configuration using both manually supplemented samples and fully automated (model-only) samples. Error bars indicate standard error. Manually derived estimates are red dashed lines, with corresponding uncertainty indicated by grey dashed lines. DPT-based configurations outperform Depth Anything in estimation accuracy; however, all configurations underestimate density, particularly under full automation.
}\label{fig:activity_chart}
\end{figure}

\textbf{Complete Pipeline -- Density and Abundance Estimation.} Population densities and abundance estimates obtained using distances derived with each computer vision
configuration for both the supplemented data (manual
frame sample) and model only data (automated frame sample) were calculated and are shown in
Figure~\ref{fig:activity_chart}. All automated input configurations underestimate density, less so when utilising manual frame selection. Under such a setting, DPT pipelines even reach close to manual estimates, well inside the uncertainty cone of CTDS generally\cite{howe2017distance}.

\textbf{Framework Statistical Effects.}\label{subsubsec:configuration_effects} The data show that the density and abundance estimates obtained using DPT are greater than those using Depth Anything.
Additionally, for each of these distance estimation models, the density and abundance estimates obtained using the bounding box detection method are also generally greater than those using segmentation detection. This gives the overall trend of a decrease in estimates from MANUAL $\rightarrow$ DPT/BBOX $\rightarrow$
DPT/SEG $\rightarrow$ DA/BBOX $\rightarrow$ DA/SEG\@. This trend can be rationalised by considering how distance influences estimated density. In essence, density estimation relies on fitting a detection function, $g(r)$, to the distribution of observed distances. This function models how detection probability decays with distance, contingent on the assumption of certain detection at zero distance ($g(0)=1$).When distances are overestimated, the fitted detection function becomes elongated, decaying at a slower rate to account for the apparent increase in detections at farther distances. In CTDS, density is estimated as
\begin{equation}
    \hat{D} = \frac{2t \sum_{k=1}^{K} n_k}{\theta w^2 \sum_{k=1}^{K} T_k \hat{P}}
\end{equation}
where $n_k$ is the count of observations at camera $k$, $T_k$ is the deployment duration, $t$ is the snapshot interval, $\theta$ is the horizontal angle of view, and $w$ is the truncation distance. The term $\hat{P}$ represents the global effective detection probability within the survey sector, calculated as 
\begin{equation}
\hat{P} = \frac{2}{w^2}\int_{0}^{w} r \, g(r)\,dr
\end{equation}
Here, $r$ denotes the radial distance from the camera; its inclusion in the integrand accounts for the increasing geometric area available at greater distances. Systematic overestimation of distances biases the fitted detection function by elongating its tail. This inflation increases the integral $\int_{0}^{w} r\, g(r)\,dr$, resulting in an overestimated $\hat{P}$. Because $\hat{P}$ appears in the denominator of the density estimator, this induces a negative bias in the final density estimate. Thus, systematic overestimation of distances—as observed in all evaluated computer vision pipelines—leads to lower density estimates. DPT/BBOX exhibited the smallest mean error ($\Delta_{\text{AVG}} = 0.586$m), while DA/SEG exhibited the largest~($\Delta_{\text{AVG}} = 2.80$m). Given the statistical relationship described above, the variation in estimates shown in Figure~\ref{fig:activity_chart} is driven by these method-specific errors.



\begin{table}[t]
\centering
\footnotesize
\caption{\textbf{Abundance and Density for Supplemented Data.}
Detailed population density and abundance estimates showing the estimate, bootstrap median, lower confidence interval (LCI), upper confidence interval (UCI), standard error (SE), and coefficient of variation (CV). Estimates using manual distances are shown for reference.}
\label{tab:supplemented_activity}

\begin{tabular}{l S[table-format=1.2,table-number-alignment=center] S[table-format=1.2,table-number-alignment=center] S[table-format=1.2,table-number-alignment=center] S[table-format=1.2,table-number-alignment=center] S[table-format=1.2,table-number-alignment=center] S[table-format=1.2,table-number-alignment=center] S[table-format=1.2,table-number-alignment=center] S[table-format=1.2,table-number-alignment=center] S[table-format=1.2,table-number-alignment=center] S[table-format=1.2,table-number-alignment=center]} \toprule \multirow{2}{*}{\textbf{Method}} & \multicolumn{5}{c}{\textbf{Density}} & \multicolumn{5}{c}{\textbf{Density Bootstrap}} \\ \cmidrule(lr){2-6} \cmidrule(lr){7-11} & {Estimate} & {LCI} & {UCI} & {SE} & {CV} & {Median} & {LCI} & {UCI} & {SE} & {CV} \\ \midrule Manual & 0.46 & 0.46 & 0.88 & 0.16 & 0.34 & 0.43 & 0.26 & 0.66 & 0.12 & 0.27 \\ DPT, BBOX & 0.36 & 0.15 & 0.87 & 0.17 & 0.48 & 0.40 & 0.21 & 0.77 & 0.16 & 0.39 \\ DPT, SEG & 0.31 & 0.16 & 0.58 & 0.10 & 0.33 & 0.31 & 0.15 & 0.56 & 0.11 & 0.36 \\ DA, BBOX & 0.23 & 0.07 & 0.73 & 0.15 & 0.65 & 0.23 & 0.10 & 0.37 & 0.08 & 0.36 \\ DA, SEG & 0.23 & 0.13 & 0.43 & 0.08 & 0.32 & 0.24 & 0.10 & 0.46 & 0.09 & 0.40 \\ \end{tabular}

\vspace{1.2em}

\begin{tabular}{l S[table-format=4.0,table-number-alignment=center] S[table-format=4.0,table-number-alignment=center] S[table-format=4.0,table-number-alignment=center] S[table-format=4.0,table-number-alignment=center] S[table-format=1.2,table-number-alignment=center] S[table-format=4.0,table-number-alignment=center] S[table-format=4.0,table-number-alignment=center] S[table-format=4.0,table-number-alignment=center] S[table-format=4.0,table-number-alignment=center] S[table-format=1.2,table-number-alignment=center]} \toprule \multirow{2}{*}{\textbf{Method}} & \multicolumn{5}{c}{\textbf{Abundance}} & \multicolumn{5}{c}{\textbf{Abundance Bootstrap}} \\ \cmidrule(lr){2-6} \cmidrule(lr){7-11} & {Estimate} & {LCI} & {UCI} & {SE} & {CV} & {Median} & {LCI} & {UCI} & {SE} & {CV} \\ \midrule Manual & 2454 & 1273 & 4730 & 841 & 0.34 & 2312 & 1202 & 4126 & 847 & 0.37 \\ DPT, BBOX & 1917 & 785 & 4683 & 920 & 0.48 & 2150 & 1241 & 3704 & 701 & 0.33 \\ DPT, SEG & 1651 & 878 & 3104 & 544 & 0.33 & 1685 & 1054 & 2748 & 512 & 0.30 \\ DA, BBOX & 1220 & 379 & 3927 & 797 & 0.65 & 1175 & 532 & 2196 & 458 & 0.39 \\ DA, SEG & 1256 & 678 & 2327 & 403 & 0.32 & 1182 & 583 & 2549 & 550 & 0.47 \\ \bottomrule \end{tabular} \vspace{-15pt}

\end{table}

\tb{\textbf{Sampling and Detector-based Effects.} As model-only data obtained from automated sampling are not associated with any manually obtained,
specific annotations in the dataset, the associated number of individual distance estimates used to estimate
density is entirely dependent on the ability of MegaDetector to identify individuals. The dense forested environment at Taï National Park inevitably results in many individuals ($36.77 \% \approx 1100$ instances) undetected in circumstances where the human annotator was indeed able to do so.
When significant occlusion is a factor, the single-frame MegaDetector is additionally handicapped by the lack of context from previous and subsequent frames that are available to the human annotator.
In addition, two individuals are in some circumstances detected as a single individual~(e.g., some infant carrying settings). While false positives do occur, they can be considered insignificant. Ultimately, these detection-based factors result in many observations being unaccounted~ (almost a thousand misses in this case) which leads to further underestimation of density and abundance. The results of the bootstrapping (see Tables~\ref{tab:supplemented_activity} and~\ref{tab:model_only_activity}) for the different configurations generally show good stability with minimal effect on the variance, error and confidence intervals, indicating that the functions fitted to the distance distributions are suitable and insensitive to random fluctuations in the data from resampling. This gives a positive insight into the density and abundance estimates and indicates that the errors are representative with respect to the distance distribution~\cite{distance_samping}}.

\vspace{-8pt}\section{Conclusions}\vspace{-6pt}
\textbf{Realistic Pathway to Automatic Great Ape Population Monitoring for Conservation.}
This study demonstrates via a realistic chimp population monitoring scenario that monocular depth estimation, when integrated with camera trap distance sampling, provides a viable pathway toward automated population estimation for great apes. Using real-world chimpanzee camera trap data, we show that automated pipelines can produce population estimates comparable to conventional manual approaches, while reducing time and labour requirements significantly. While our reported performance approaches that of manual methods, systematic biases and detection failures remain limiting factors. Addressing these through improved calibration, more robust localisation, and depth estimation models tailored to complex forest environments will be essential for operational deployment. Overall, these results provide a clear proof of concept and indicate strong potential for scalable, automated bio-monitoring to support large-scale biodiversity assessment for great apes and beyond.
\begin{table}[!h]
\centering
\footnotesize
\caption{\textbf{Abundance and Density for Model-Only Data.}
Population estimates obtained using model-derived distances only. Reported statistics follow the same convention as Table~\ref{tab:supplemented_activity}.}
\label{tab:model_only_activity}

\begin{tabular}{l
S[table-format=1.2] S[table-format=1.2] S[table-format=1.2] S[table-format=1.2] S[table-format=1.2]
S[table-format=1.2] S[table-format=1.2] S[table-format=1.2] S[table-format=1.2] S[table-format=1.2]}
\toprule
\multirow{2}{*}{\textbf{Method}} & \multicolumn{5}{c}{\textbf{Density}} & \multicolumn{5}{c}{\textbf{Density Bootstrap}} \\
\cmidrule(lr){2-6} \cmidrule(lr){7-11}
& {Est.} & {LCI} & {UCI} & {SE} & {CV}
& {Median} & {LCI} & {UCI} & {SE} & {CV} \\
\midrule
Manual     & 0.46 & 0.46 & 0.88 & 0.16 & 0.34 & 0.43 & 0.26 & 0.66 & 0.12 & 0.27 \\
DPT, BBOX  & 0.36 & 0.15 & 0.87 & 0.17 & 0.48 & 0.40 & 0.21 & 0.77 & 0.16 & 0.39 \\
DPT, SEG   & 0.31 & 0.16 & 0.58 & 0.10 & 0.33 & 0.31 & 0.15 & 0.56 & 0.11 & 0.36 \\
DA, BBOX   & 0.23 & 0.07 & 0.73 & 0.15 & 0.65 & 0.23 & 0.10 & 0.37 & 0.08 & 0.36 \\
DA, SEG    & 0.23 & 0.13 & 0.43 & 0.08 & 0.32 & 0.24 & 0.10 & 0.46 & 0.09 & 0.40 \\
\bottomrule
\end{tabular}

\vspace{1.2em}

\begin{tabular}{l
S[table-format=4.0] S[table-format=4.0] S[table-format=4.0] S[table-format=4.0] S[table-format=1.2]
S[table-format=4.0] S[table-format=4.0] S[table-format=4.0] S[table-format=4.0] S[table-format=1.2]}
\toprule
\multirow{2}{*}{\textbf{Method}} & \multicolumn{5}{c}{\textbf{Abundance}} & \multicolumn{5}{c}{\textbf{Abundance Bootstrap}} \\
\cmidrule(lr){2-6} \cmidrule(lr){7-11}
& {Est.} & {LCI} & {UCI} & {SE} & {CV}
& {Median} & {LCI} & {UCI} & {SE} & {CV} \\
\midrule
Manual     & 2454 & 1273 & 4730 & 841 & 0.34 & 2312 & 1202 & 4126 & 847 & 0.37 \\
DPT, BBOX  & 1917 &  785 & 4683 & 920 & 0.48 & 2150 & 1241 & 3704 & 701 & 0.33 \\
DPT, SEG   & 1651 &  878 & 3104 & 544 & 0.33 & 1685 & 1054 & 2748 & 512 & 0.30 \\
DA, BBOX   & 1220 &  379 & 3927 & 797 & 0.65 & 1175 &  532 & 2196 & 458 & 0.39 \\
DA, SEG    & 1256 &  678 & 2327 & 403 & 0.32 & 1182 &  583 & 2549 & 550 & 0.47 \\
\bottomrule
\end{tabular}\vspace{-5pt}

\end{table}
\newpage

%
%
%
\bibliographystyle{splncs04}
\bibliography{mybibliography}

\end{document}